\documentclass[letterpaper]{article} 
\usepackage{aaai25}  
\usepackage{times}  
\usepackage{helvet}  
\usepackage{courier}  
\usepackage[hyphens]{url}  
\usepackage{graphicx} 
\usepackage{natbib}  
\usepackage{caption} 
\usepackage{algorithm}
\usepackage{algorithmic}
\usepackage{multirow}
\usepackage{subcaption}

\usepackage{newfloat}
\usepackage{listings}
\DeclareCaptionStyle{ruled}{labelfont=normalfont,labelsep=colon,strut=off} 
\floatstyle{ruled}
\newfloat{listing}{tb}{lst}{}
\floatname{listing}{Listing}
\title{Planning with Vision-Language Models and a 
Use Case in Robot-Assisted Teaching}
 
\author {
    Xuzhe Dang\textsuperscript{\rm 1},
    Lada Kudláčková\textsuperscript{\rm 2},
    Stefan Edelkamp\textsuperscript{\rm 1,2}
}
\affiliations {
    \textsuperscript{\rm 1}
    Computer Science Department, Artificial Intelligence Center,
Faculty of Electrical Engineering, Czech Technical University, Prague, Czech Republic\\
    \textsuperscript{\rm 2}Department of Theoretical Computer Science and Mathematical Logic,
Faculty of Mathematics and Physics, Charles University, Prague, Czech Republic\\
    dangxuzh@fel.cvut.cz, kudlackova@ktiml.cuni.cz, stefan.edelkamp@gmail.com
}

\usepackage{bibentry}

\begin{document}

\maketitle

\begin{abstract}
Automating the generation of Planning Domain Definition Language (PDDL) with Large Language Model (LLM) opens new research topic in AI planning, particularly for complex real-world tasks. This paper introduces Image2PDDL, a novel framework that leverages Vision-Language Models (VLMs) to automatically convert images of initial states and descriptions of goal states into PDDL problems. By providing a PDDL domain alongside visual inputs, Imasge2PDDL addresses key challenges in bridging perceptual understanding with symbolic planning, reducing the expertise required to create structured problem instances, and improving scalability across tasks of varying complexity. We evaluate the framework on various domains, including standard planning domains like blocksworld and sliding tile puzzles, using datasets with multiple difficulty levels. Performance is assessed on syntax correctness, ensuring grammar and executability, and content correctness, verifying accurate state representation in generated PDDL problems. The proposed approach demonstrates promising results across diverse task complexities, suggesting its potential for broader applications in AI planning.
We will discuss a potential use case in robot-assisted teaching of students with Autism Spectrum Disorder. 
\end{abstract}

%

\section{Introduction}
Automating the generation of Planning Domain Definition Language (PDDL) problems has long been a challenge in AI planning, particularly for applications involving complex, real-world tasks. Traditional approaches to PDDL generation often require domain-specific knowledge and substantial manual effort to accurately structure problem instances. This barrier limits the scalability and accessibility of AI planning, as defining object relationships, spatial configurations, and task goals is labor-intensive and demands expertise.

Recent advances in Large Language Models (LLMs) have opened new possibilities by enabling models to interpret and translate visual and textual data into structured, symbolic formats, bridging the gap between perceptual understanding and symbolic reasoning \cite{liu2023llm+, xie2023translating, shirai2024vision}. However, each approach has its limitations. LLM+P \cite{liu2023llm+} requires text descriptions for both initial and goal states, which, in practical applications, depends on human input or additional tools to convert images to text. Xie’s work \cite{xie2023translating} focuses solely on translating text descriptions of goal states, lacking a mechanism to interpret visual data. Meanwhile, ViLaIn \cite{shirai2024vision} can generate PDDL problems from images but relies on an object detection model and a captioning model to process images, adding complexity to the pipeline. These limitations highlight the need for a more streamlined approach that can directly process both visual and textual inputs for automated PDDL problem generation.

To address these challenges, we present Image2PDDL, a novel framework that leverages Vision-Language Models (VLMs) to automatically generate PDDL problems from both visual and textual inputs. Image2PDDL is designed to handle a wide range of input formats, including images of initial and goal states as well as textual descriptions, allowing it to adapt seamlessly across diverse domains and task complexities. By directly interpreting spatial and categorical relationships within visual data and integrating them with goal-oriented descriptions, Image2PDDL provides a flexible solution for PDDL problem generation. We evaluated the framework across traditional planning domains—Blocksworld and Sliding-Tile Puzzle—as well as a 3D world domain, Kitchen, assessing both syntax correctness and content correctness of the generated PDDL problems. In all domains, Image2PDDL demonstrated promising results, effectively reducing the need for domain-specific expertise and making AI planning more accessible, scalable, and applicable to real-world scenarios.

Image2PDDL makes several key contributions to the field of AI planning. Its adaptability across both traditional planning domains and complex 3D environments underscores its versatility and scalability, positioning it as a valuable tool for a wide range of AI planning applications. By reducing dependency on domain-specific knowledge, Image2PDDL broadens accessibility and opens new possibilities for applying AI planning to real-world tasks. Future work could focus on enhancing the model’s capacity to interpret more intricate object relationships and dynamic scenarios, further expanding the potential of automated planning across diverse and complex environments.

\begin{figure*}[h]
    \centering
    \includegraphics[width=0.8\textwidth]{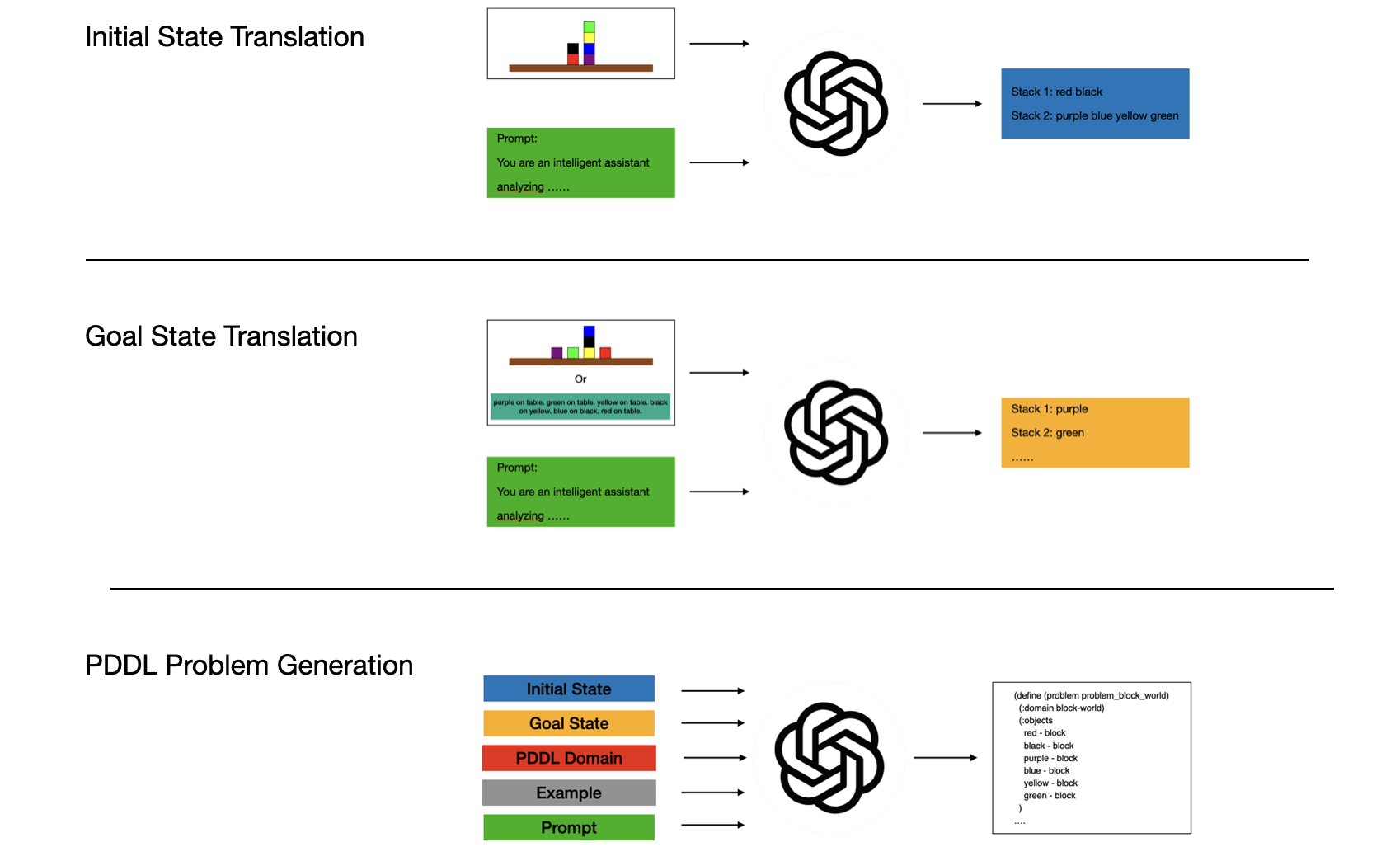}
    \caption{Image2PDDL operates in three main steps. First, the image of the initial state is translated into a predefined state format. Next, either an image or text description of the goal state is similarly converted to the same state format. Finally, both states are used to generate a PDDL problem based on predefined domains and examples.}
    \label{fig:method}
\end{figure*}

We will first introduce Image2PDDL together with experiments in several benchmark domains, and then discuss a
case in automated teaching of students with Autism Spectrum Disorders (ASD).

\section{Related Work}

Recent advances in LLMs and VLMs have spurred interest in their applications to automated planning. Most current work in this area focuses on using LLMs and VLMs as planners, where the models are tasked with generating plans directly based on high-level descriptions or visual inputs \cite{huang2022language, raman2022planning, zhang2023grounding, dagan2023dynamic, guan2023leveraging}. In these approaches, the models produce a sequence of symbolic actions \cite{lin2023text2motion}, code \cite{liang2023code}, or predefined skills \cite{singh2023progprompt, brohan2023can} to complete a task, acting as planners that bridge the gap between high-level goals and executable actions. A significant drawback of these methods is that the generated plans are not always guaranteed to be correct, as both LLMs and VLMs may produce sequences that overlook domain constraints or violate task feasibility. Additionally, the planning process is inherently opaque, making it difficult to verify or explain the rationale behind the generated plans.

Recent research has also explored the potential of LLMs and VLMs in directly generating PDDL domains or PDDL problems, shifting the focus from plan generation to the formulation of structured problem instances \cite{liu2023llm+, xie2023translating, shirai2024vision, smirnov2024generating, oswald2024large}. These approaches aim to leverage the language and visual understanding capabilities of LLMs and VLMs to automate the creation of PDDL problem definitions, which include initial and goal states and domain-specific constraints.

\section{Method}
The Image2PDDL framework leverages VLM to parse visual and textual inputs, transforming them into Planning Domain Definition Language (PDDL) problem representations based on pre-defined domains. The core idea is to utilize the spatial relationship understanding that VLMs offer, enabling the translation of complex images and descriptions into structured PDDL problems, which traditionally require extensive domain expertise to manually create.

\subsection{Framework Overview}
Image2PDDL consists of a pipeline, as shown in Figure \ref{fig:method}, that translates an image of an initial state and either an image or text description of a goal state into PDDL problem format. This is achieved in in the following three main steps, each using ChatGPT4o for structured output generation.

\textbf{Initial State Translation}: First, we input the image of the initial state into ChatGPT4o, which is guided by a structured prompt. This prompt includes an example of the desired output format to encourage accurate parsing of spatial relationships between objects. The model generates a pre-defined textual representation of the spatial relationships observed in the initial state, such as object locations and their relative positions.

\textbf{Goal State Translation}: For the goal state, we offer either an image or a text description. Similarly to the initial state, ChatGPT4o is prompted to parse this goal state information into the same structured format, ensuring consistency between the initial and goal state representations. This format standardization is crucial for correctly translating the data into PDDL syntax.

\begin{table*}[h]
\centering
\begin{tabular}{|l|p{6cm}|p{4cm}|p{4cm}|}
\hline
\textbf{Domain}           & \textbf{Object Types}                                                                                 & \textbf{Predicates}                                    & \textbf{Actions}                                  \\
\hline
Blocksworld     & \texttt{block}                                                                                         & \texttt{on}, \texttt{ontable}, \texttt{clear}, \texttt{holding}, \texttt{arm-empty} & \texttt{pick-up}, \texttt{put-down}, \texttt{stack}, \texttt{unstack} \\
\hline
Sliding-Tile Puzzle & \texttt{tile}, \texttt{position}                                                                              & \texttt{tile}, \texttt{position}, \texttt{at}, \texttt{blank}, \texttt{inc}, \texttt{dec} & \texttt{move-down}, \texttt{move-up}, \texttt{move-right}, \texttt{move-left} \\
\hline

Kitchen          & \texttt{item} (\texttt{fruit}: \texttt{lemon}, \texttt{apple}; \texttt{cutting\_board}; \texttt{mug}: \texttt{black\_mug}, \texttt{green\_mug}; 
\texttt{kettle}; 
\texttt{wine}; 
\texttt{soda}: 
\texttt{cola}, 
\texttt{fanta}), 
\texttt{location} (\texttt{counter}, \texttt{sink}, \texttt{shelf}, \texttt{stove}) & \texttt{at}, \texttt{clear} & \texttt{move-item}                               \\
\hline
\end{tabular}
\caption{Defined Object Types, Predicates, and Actions in the Domain Descriptions}
\label{tab:domain_descriptions}
\end{table*}

\begin{figure}[t]
    \centering
    \includegraphics[width=0.5\textwidth]{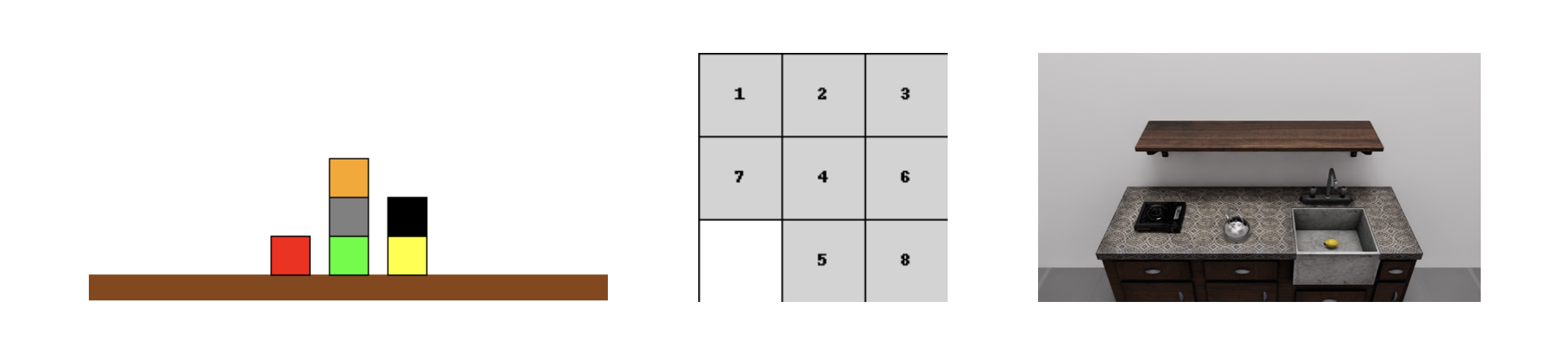}
    \caption{Images of example scenarios of each domain.}
    \label{fig:example}
\end{figure}

\textbf{PDDL Problem Generation}: With structured text representations of both the initial and goal states, we proceed to PDDL problem generation. ChatGPT4o receives the initial and goal states, the corresponding PDDL domain definition, and an example format to guide output structure. Additional information, such as predicate rules and detailed state structure definitions, is also included to refine output quality and enhance syntax correctness and executability.

This pipeline allows Image2PDDL to leverage the spatial reasoning capabilities of VLMs, providing an automated, scalable solution to generate PDDL problems for diverse planning domains. The iterative prompt refinements and examples enable ChatGPT4o to achieve high accuracy in syntax and content representation, making the framework effective for complex task definitions across different difficulty levels.

\section{Experiments and Evaluation}

\subsection{Dataset}
To evaluate the Image2PDDL framework, we prepared datasets across three distinct domains: Blocksworld, Sliding-Tile Puzzle, and Kitchen Domain. Each domain includes scenarios categorized by difficulty levels—Easy, Medium, and Hard—allowing us to test the framework’s scalability and accuracy across a range of task complexities. For each difficulty level, we prepared 50 unique scenarios. Table \ref{tab:domain_descriptions}  shows object types, predicates, and actions of each domains. Figure \ref{fig:example} shows image of an scenario from each domain.

The Blocksworld domain assesses Image2PDDL’s ability to interpret spatial relationships by arranging blocks, with difficulties based on the number of blocks (5 for Easy, 6 for Medium, and 7 for Hard). In the Sliding-Tile Puzzle domain, the framework rearranges tiles to match a goal configuration, testing sequential spatial handling across three puzzle sizes: 8 tiles (Easy), 15 tiles (Medium), and 24 tiles (Hard). The Kitchen Domain, created in IsaacSim, evaluates real-world scenario comprehension by identifying and categorizing objects. Task difficulty increases from identifying a single item’s location (Easy) to distinguishing between two items by attributes (Medium) and recognizing three items with finer detail, such as brand or color (Hard).

For each scenario, the dataset provides an initial and goal image alongside a textual goal state description, enabling Image2PDDL to process diverse data formats and effectively translate them into structured PDDL problems. This setup allows for a systematic evaluation of the framework’s ability to handle different levels of task complexity across varied planning domains.

\subsection{Results}

Performance was assessed based on two metrics: Syntax Correctness and Content Correctness. Syntax correctness was verified by passing the generated PDDL problems to a Fast Downward planner \cite{helmert2006fast} to ensure each problem could be successfully parsed and executed without syntax errors. Content correctness was evaluated by comparing the initial and goal states described in the generated PDDL problem against the true states, verifying accurate representation of object locations and relationships. To examine the flexibility of Image2PDDL, we used both images and text descriptions to represent goal states, analyzing the framework’s ability to interpret both visual and textual inputs accurately. Tables \ref{tab:block_world_errors}, \ref{tab:slide_tile_puzzle_errors}, and \ref{tab:kitchen_domain_errors} provide detailed error counts across domains and difficulty levels, showcasing the method’s performance across varying task complexities.

\textbf{Syntax Correctness}: Overall, Image2PDDL demonstrated strong syntax correctness across all domains, with zero syntax errors recorded across both input modalities—images for both initial and goal states, and images for the initial state paired with a text description for the goal state. The generated PDDL problems were consistently verified as syntactically valid by the Fast Downward planner, showing that the framework adheres effectively to PDDL syntax rules. Occasionally, the VLM returned PDDL problems with syntax highlighting, but these highlights were easily removed without affecting the problem structure. This consistency suggests that Image2PDDL reliably constructs grammatically correct and executable PDDL statements, regardless of input type or task complexity within each domain.

\begin{table}[h]
\centering
\resizebox{0.49\textwidth}{!}{ 
\begin{tabular}{|l|c|c|c|}
\hline
\textbf{Input Type}                       & \textbf{Difficulty} & \textbf{Syntax Errors} & \textbf{Content Errors} \\
\hline
\multirow{3}{*}{Images (Initial \& Goal)} & Easy                & 0                      & 1/50                       \\
                                          & Medium              & 0                      & 3/50                       \\
                                          & Hard                & 0                      & 3/50                       \\
\hline
\multirow{3}{*}{Image (Initial), Text (Goal)} & Easy           & 0                      & 1/50                       \\
                                             & Medium         & 0                      & 3/50                       \\
                                             & Hard           & 0                      & 3/50                       \\
\hline
\end{tabular}
}
\caption{Syntax and Content Errors in Blocksworld Domain}
\label{tab:block_world_errors}
\end{table}

\begin{table}[h]
\centering
\resizebox{0.49\textwidth}{!}{ 
\begin{tabular}{|l|c|c|c|}
\hline
\textbf{Input Type}                       & \textbf{Difficulty} & \textbf{Syntax Errors} & \textbf{Content Errors} \\
\hline
\multirow{3}{*}{Images (Initial \& Goal)} & Easy                & 0                      & 2/50                       \\
                                          & Medium              & 0                      & 1/50                       \\
                                          & Hard                & 0                      & 1/50                       \\
\hline
\multirow{3}{*}{Image (Initial), Text (Goal)} & Easy           & 0                      & 2/50                       \\
                                             & Medium         & 0                      & 1/50                       \\
                                             & Hard           & 0                      & 2/50                       \\
\hline
\end{tabular}
}
\caption{Syntax and Content Errors in Sliding-Tile Puzzle Domain}
\label{tab:slide_tile_puzzle_errors}
\end{table}

\begin{table}[h]
\centering
\resizebox{0.49\textwidth}{!}{ 
\begin{tabular}{|l|c|c|c|c|}
\hline
\textbf{Input Type}                       & \textbf{Difficulty} & \textbf{Syntax Errors} & \textbf{Content Errors} \\
\hline
\multirow{3}{*}{Images (Initial \& Goal)} & Easy                & 0                      & 7/50                       \\
                                          & Medium              & 0                      & 5/50                       \\
                                          & Hard                & 0                      & 9/50                       \\
\hline
\multirow{3}{*}{Image (Initial), Text (Goal)} & Easy           & 0                      & 3/50                       \\
                                             & Medium         & 0                      & 3/50                       \\
                                             & Hard           & 0                      & 3/50                       \\
\hline
\end{tabular}
}
\caption{Syntax and Content Errors in Kitchen Domain}
\label{tab:kitchen_domain_errors}
\end{table}

\begin{figure}[h!]
    \centering
    \begin{minipage}{0.2\textwidth}
        \centering
        \includegraphics[width=\textwidth]{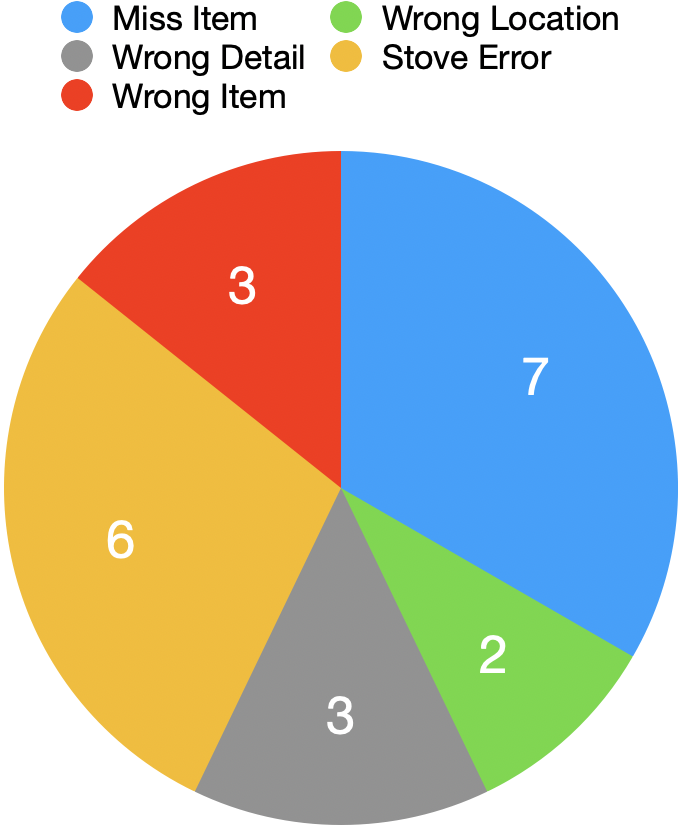}
    \end{minipage}\hfill
    \begin{minipage}{0.2\textwidth}
        \centering
        \includegraphics[width=\textwidth]{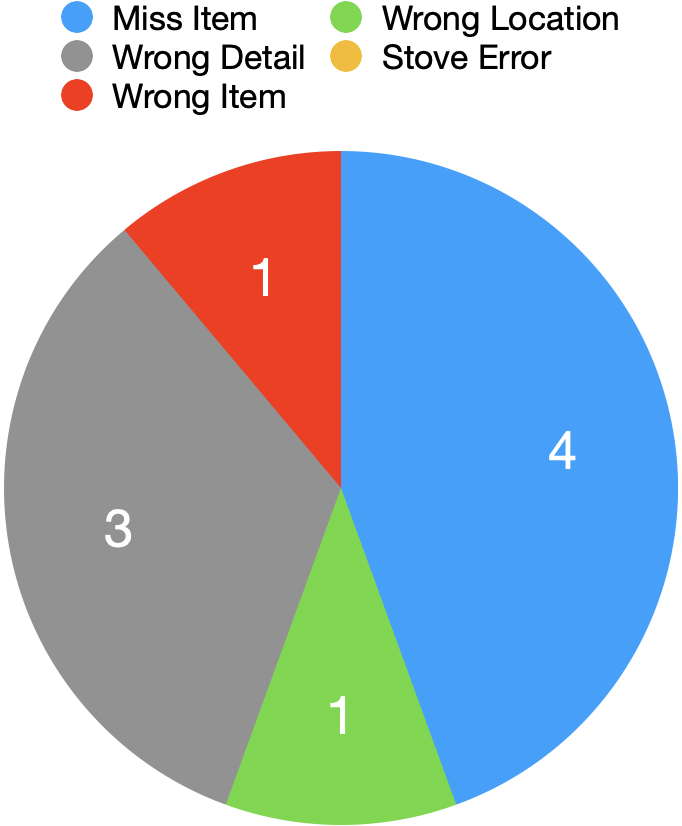}
    \end{minipage}
    \caption{Types of Errors in the Kitchen Domain: The left chart represents errors when using an image to describe the goal state, while the right chart represents errors when using a text to describe the goal state.}
    \label{fig:error_type}
\end{figure}

\textbf{Content Correctness}: Our analysis shows that as domain complexity increases, Image2PDDL is more prone to generating incorrect states in PDDL problems. We reviewed errors within each domain to identify recurring patterns and domain-specific challenges.

In the Blocksworld domain, Image2PDDL consistently recognized the correct number of objects across both image and text-based goal states, demonstrating reliable object identification. However, the same errors appeared across scenarios, revealing a recurring issue in goal state interpretation. These errors typically involved incorrect stacking orders: blocks intended to be on the table were sometimes placed atop stacks, while blocks meant to be on top were placed on the table. This reversal in stacking sequences indicates a systematic misinterpretation of positional relationships in the generated PDDL problem.

In the Sliding-Tile Puzzle domain, errors primarily involved switched tile locations, where specific tiles were incorrectly swapped. Although Image2PDDL generally captured the spatial configuration well, it occasionally struggled with precise tile placement, particularly in more complex tile arrangements.

In the Kitchen domain, the visually intricate environment introduced additional sources of error. We classified these errors into five categories: (1) Stove Error, where the LLM mistakenly identifies the stove as an item instead of a location; (2) Missing Item, where certain items are omitted from the generated PDDL problem; (3) Wrong Item, where items are misclassified, leading to inaccuracies in object identification; (4) Wrong Detail, where items are assigned incorrect attributes (e.g., brand or color); and (5) Wrong Location, where items are placed in incorrect positions. Figure \ref{fig:error_type} presents the distribution of these error types. Across goal state representation formats, the most common error was missing items, typically smaller objects like fruit or soda cans. Additionally, using images to represent the goal state often led to stove misidentifications, where the stove was treated as an item rather than a location, a recurring issue when defining goal states.

\section{Image2PDDL Use Case}

Robot-assisted teaching of students with Autism Spectrum Disorder
Teachers utilize different types of prompts to support student learning. In robot-assisted education, robots can deliver physical, verbal, visual or gestural prompts, and can teach problem-solving through demonstration. Research indicates that students with Autism Spectrum Disorder (ASD) often respond
positively to interactions with robots~\cite{lada1}.

Robot-assisted teaching shows potential for significant enhancements in educational outcomes of students with ASD. Educators note that children with ASD may benefit from 
robot predictability and consistency, especially with humanoid robots. However, further research is needed to verify these
observations. It is essential to identify specific use cases and conditions under which skills acquired during
robot-assisted sessions can be effectively transferred to the child’s daily life. Additionally, we should
consider the effects of robot morphology on children's responses and behavior. Collaboration with ASD
experts is crucial for developing systems that are both effective and appropriate for use in special education~\cite{lada2}. Students with ASD often show a strong interest in technological devices, including robots. 

The Codey Rocky robot is an educational tool designed to teach programming through block-based and text-based code. Studies involving this robot indicate that students with ASD are motivated to engage in programming activities, and that robot-mediated interventions can enhance peer cooperation~\cite{lada3}. Robots can also support the development of communication skills and help reduce maladaptive behaviors in students
with ASD~\cite{lada4}. Humanoid robots, such as NAO, have shown substantial benefits in teaching children with
ASD and supporting the acquisition of new skills. Focus and attention span were improved in robot-assisted
activities~\cite{lada5}.

Some researchers focus on challenges specific to individuals with ASD, such as joint attention
and imitation skills, utilizing methods based on Computer-Assisted Therapies (CATs) and Robot-Assisted
Therapies (RATs) to improve student skills~\cite{lada6}. However, some contradictory results were published on the
effects of robots on joint attention~\cite{lada7} and the overall skill development. For instance, while some students
found interactions with robots engaging, they did not demonstrate significant learning progress~\cite{lada8}.
Earlier studies have emphasized the use of video models to improve fine and gross motor task
perfor-mance in individuals with ASD~\cite{lada9}. Mobile applications have also shown potential as effective tools
for students with ASD; for example, a mathematical problem-solving app was well received by participants,
who demonstrated the ability to complete tasks independently~\cite{lada10}.

Recently, interventions based on virtual reality have been investigated to explore effects on
enhancing social skills of children~\cite{lada11} and to support remote team collaboration of autistic adults~\cite{lada12}. In an
industrial context, experiments examining the task of robotic collaborative assembly have implied that some
persons with ASD can collaborate very effectively with the robot. However, behavioral differences in gaze,
gestures, etc. suggest that solutions designed for neurotypical participants may not align with the needs of the
ASD groups~\cite{lada13}.

\subsection{TEACCH® Autism Program}

The TEACCH® Autism Program was developed at the University of North Carolina ~\cite{lada14}. It was created in the 1970s by Eric Schopler who disproved the hypothesis that autism is a mental illness caused by emotionally cold parents. His experiments imply that autism is not primarily a disorder of emotions but a disorder of processing sensory information~\cite{lada15}.

\begin{figure}
    \centering
    \includegraphics[width=0.9\linewidth]{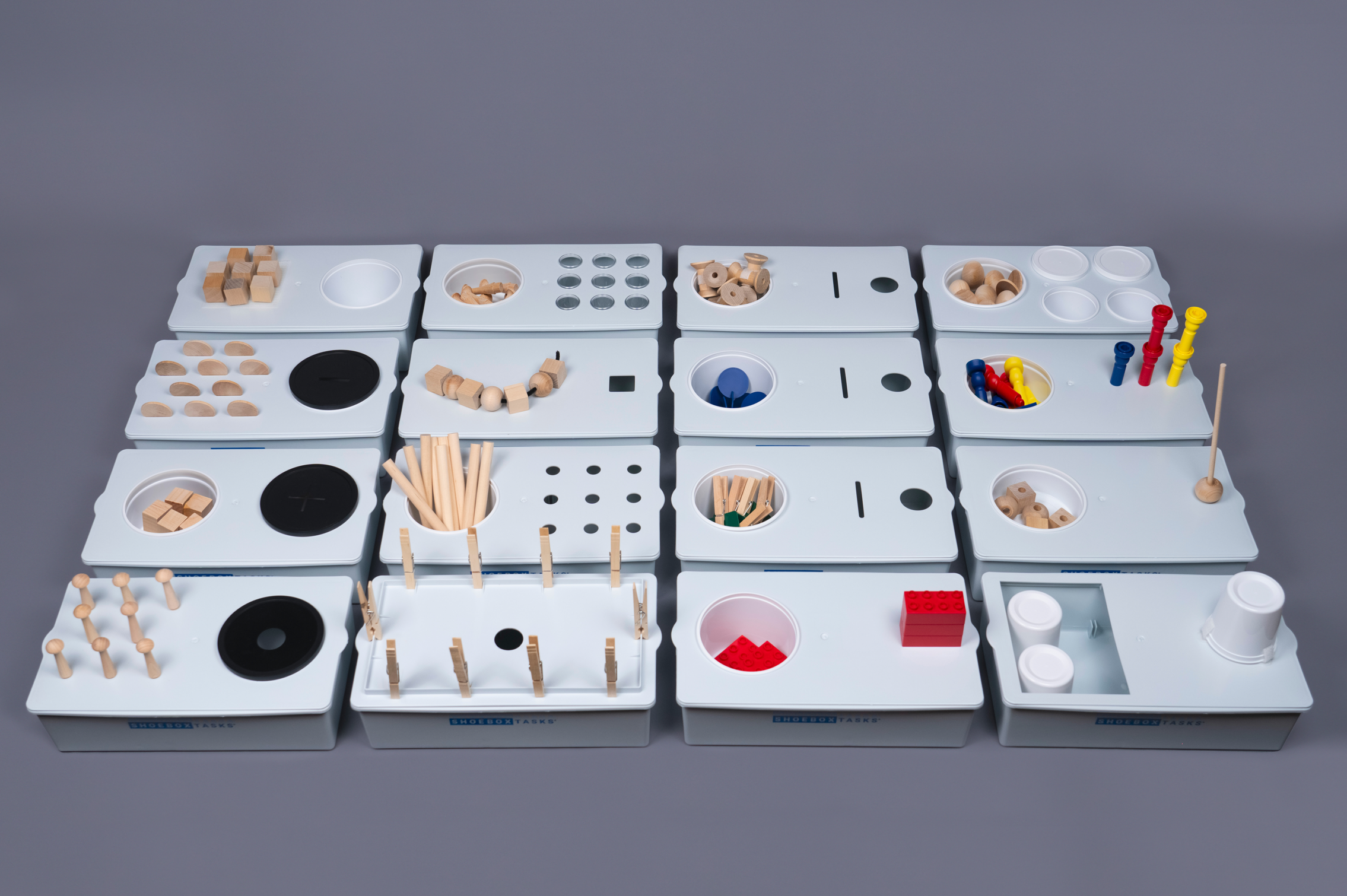}
    \caption{Shoebox tasks as a use case for robot-assisted teaching.}
    \label{fig:shoebox}
\end{figure}

The TEACCH approach exploits the premise that people with ASD are predominantly visual learners and prefer visually cued instructions in the educational process~\cite{lada16}. It's the recommended methodology in the Czech Republic for the education of people with ASD and is used worldwide. The TEACCH
Structured Work Session consists of structured tasks. Educators often use a specific type of structured task,
the shoe-box task (see Figure~\ref{fig:shoebox}), 
to be solved by students in the class.

\subsection {Automated assessment and planning of shoe-box tasks}
There is a visual structure that allows to recognize how the shoe-box task should be completed. Vision-based machine learning methods seems to be suitable for exploring tasks and automatically evaluating them. A system that provides automatic assessment of structured tasks extended by actions planning could be used in structured teaching, with a real robot in the role of teacher. To the best of our knowledge, there is no existing research on automated assessment of shoe-box tasks, or on automated action planning for robot-assisted shoe-box activities.

For learning new shoe-box tasks, students typically follow teacher guidance, or use picture-based or text-based instructional systems~\cite{lada17}. Additional experiments have explored alternative types of prompts, including video, computer, and other devices. For example, a Personal Digital Assistant was used to assist students in completing new tasks and to facilitate smoother transitions between tasks~\cite{lada18}.

Some research addresses action planning for structured tasks (not for shoe-box tasks), employing a mathematical framework with an agent in the role of a teacher. In experiments, a humanoid NAO robot was utilized in a storytelling scenario to guide the child’s gaze toward target screens, using a sequence of planned actions~\cite{lada19,lada20}.
 
Experiments with real robots can be performed with the humanoid robots and robot manipulators. Depth cameras can be used as main sensors to interpret the environment. 

\subsection{Objective}

The main goal of a use case for Image2PDDL is to implement a computerized system for the automatic assessment and planning of structured shoe-box tasks in robot-assisted education.

This system provides methods for creating models of the shoe-box tasks, automatically identifying the current state of a task instance and planning actions for it. It allows a robot to assist a student in completing the shoe-box task, which will be demonstrated by the real robot. The robot responds to the student's requests for help and provides instructions to complete the task. There are implemented multiple interaction strategies that represent different teaching approaches.
Each strategy contains, for each task instance, a sequence of robot actions that lead to successful completion of the task, if the student follows robot instructions.

\subsection{Methods of Research}

The TEACCH Structured Work Session consists of \emph{structured tasks}. The shoebox task is a structured task with a box that serves as a workstation and contains all materials for the task (puzzles, cards, blocks, \ldots). There is a visual structure that indicates how the task should be completed.

Structured task assessment are answers to the following questions: What is the goal? Was the goal achieved in this instance? What needs to be done to achieve the goal?

We can represent the structured task as abstract structure containing set of given elements with pre-defined states, actions and goal. For this structure, logic programming could be used to search for sequence of actions (plan) to reach the goal. Then, for a given task instance (i.e., configuration of elements in the box), the vision-based method Image2PDDL can be used to decide whether the task has been completed or what state it is in.

In the structured teaching sessions, there is typically a teacher that supervises a student while he is working on the task. System for automated task assessment could be used as a tool to provide basic feedback to a student, for example to confirm that the task is completed. The teacher also assists the student by giving prompts or by demonstration. This part of the teaching process can be performed by a real robot that would help the student to solve the task, using the action plan to reach the goal from current state.

Another approach might be to try to infer the abstract structure and to create a logic model automatically using machine learning methods, e.g., from demonstration (video) or from annotated samples of completed and incomplete task instances (images, simulation snapshots). Planning would use some predefined actions to manipulate objects.

Ideally, the system would not require any programming skills to be used: for a new task, the teacher creates the model just by demonstration of the task. Then, for individual students and task instances, s/he choose an educational strategy which would be performed by the robot.

\subsubsection{Representation of shoe-box tasks}

Each task type is represented by its model that contains elements, locations, initial state, goal states and possible actions. This PDDL model can be provided with the shoe-box task, or created automatically using vision-based machine learning.
For each task instance the Image2PDDL model then allows to identify automatically (from camera data) the current state
of the instance and to plan actions for the current state.

\begin{figure}
    \centering
    \includegraphics[width=0.5\linewidth]{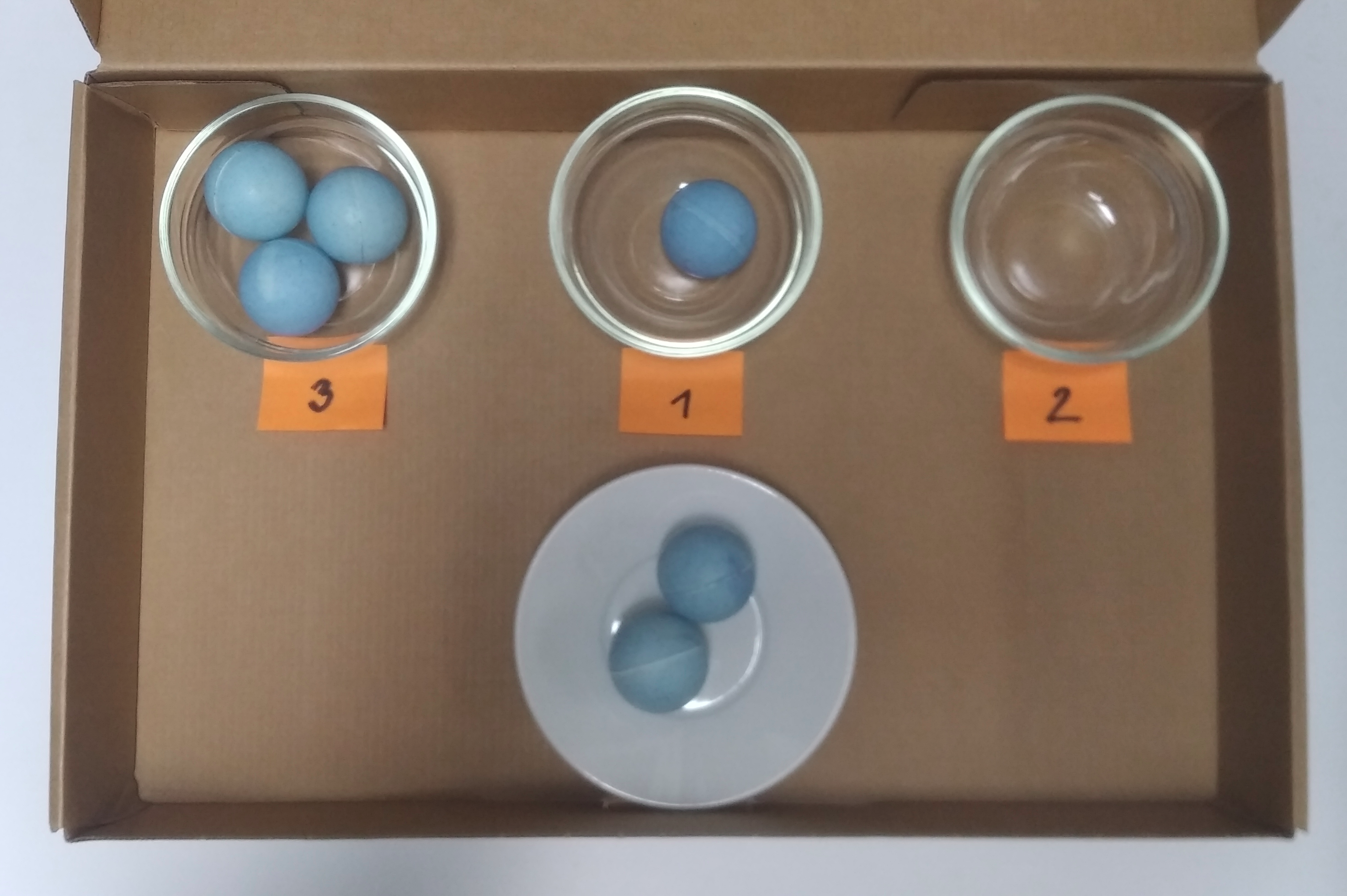}
    \caption{Instance of the distribution task that requires single snapshot to obtain specification from ChatGPT4o.}
    \label{fig:shoebox_312}
\end{figure}

\begin{figure}
    \centering
    \includegraphics[width=0.5\linewidth]{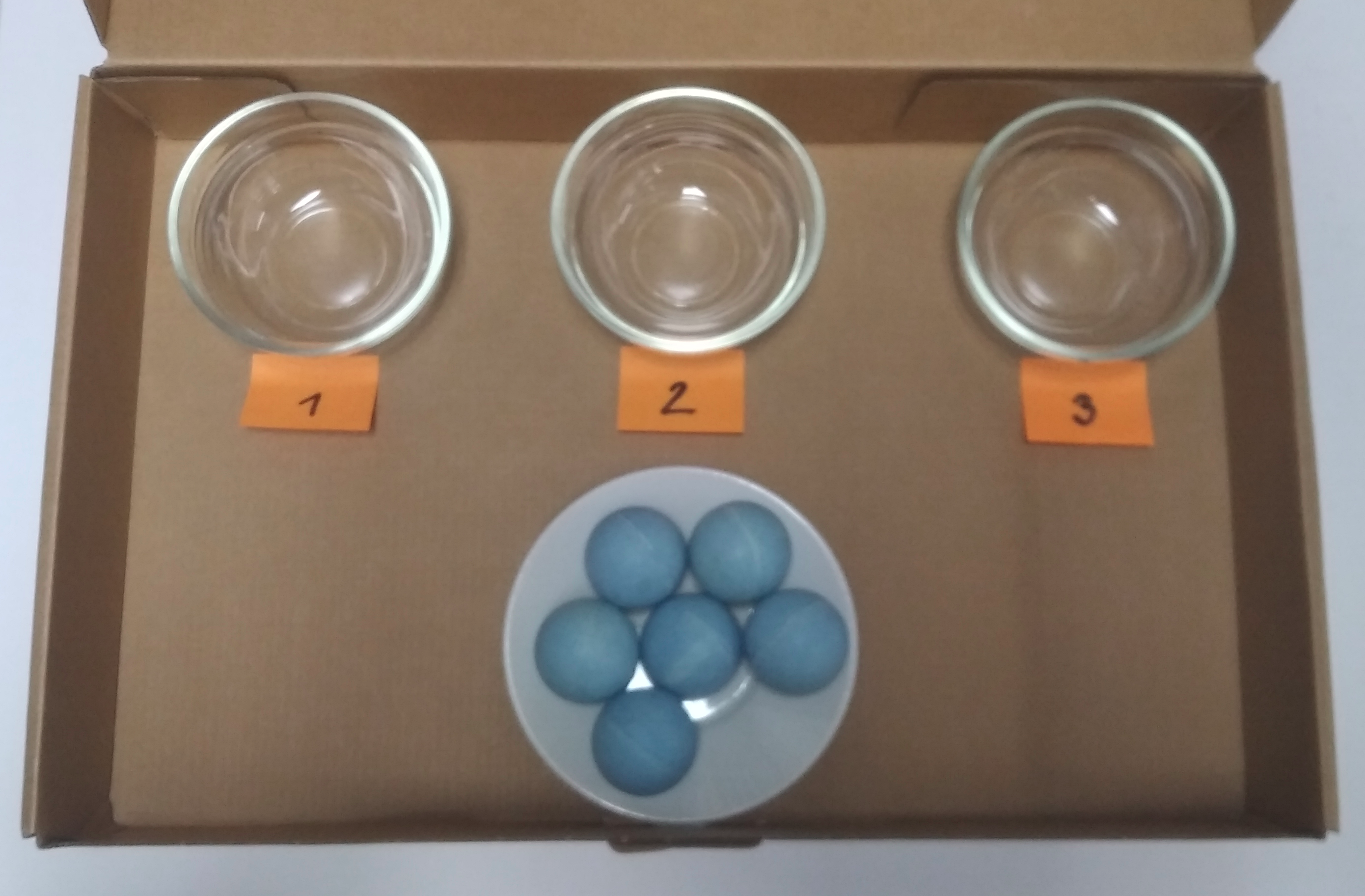}
    \caption{Instance of the distribution task with unambiguous assignment: labels are treated as bowl names, not as number of balls.}
    \label{fig:shoebox_123}
\end{figure}

\subsection{Initial experiments}

We conducted tests with a small dataset of shoebox tasks to verify that Image2PDDL can be used to transform visual and textual input into PDDL problem representations and identify shoe-box task states.

For simple tasks such as distributing balls into labeled bowls, ChatGPT4o can easily derive the goal of the task and also plan actions. In some cases, a single snapshot is sufficient to obtain a full task description. However, we usually need to provide snapshots of multiple configurations to avoid ambiguous conclusions.

\subsubsection{One-to-one correspondence task}
For the initial experiments, we chose a specific type of shoebox task: a simple put-in task with a one-to-one correspondence between objects and their target locations. We created a description of the PDDL domain and an example of the required output files, common to all scenarios in this domain.

The datasets were categorized by input data format.
Each scenario in the ShoeboxImage category contains one input image (photo or diagram) depicting a specific task configuration.
The scenarios in the ShoeboxVideoSnapshots category are represented by two video snapshots of different task states, one of which is a snapshot of the initial state.

One of the main challenges in the analysis of visual input is objects that are partially obscured or not visible at all. We improved the generated task description by extending the Image2PDDL with simple action planning. The objects that are necessary for solving the shoe-box task are involved in at least one action and would be recognized, even if they were not identified in the images.
\\

\textbf{Syntax Correctness}:
The syntax correctness of the PDDL problems was verified using a Fast Downward Planner. For both categories, the generated PDDL problems were consistently valid.
\\

\textbf{Content Correctness}:
In some scenarios with multiple items of the same type, the number of items in the generated text description and PDDL problem was incorrect, even though all items were visible in the image. In the ShoeboxVideoSnapshots category, we prompt ChatGPT4o to analyze the items in each snapshot separately, compare the results, and determine the total list of task objects. This comparison increased the accuracy of the number of objects identified.

In a few cases, some objects were misclassified or not recognized at all.

\subsubsection{Robot as a teacher}

The robot will assist the student in completing the shoe-box task e.g., by manually moving elements, pointing, or giving verbal prompts. The student can ask the robot for help, e.g., by using a button, gesture, or voice commands. 

There are many interaction strategies to decide what the robot should do and when. We will consult the strategies with experts in ASD and implement multiple recommended strategies, as options for the user (teacher). Planned actions will be translated into abstract robot actions using the selected strategy; abstract
robot actions will be translated into commands for the real robot.

The Image2PDDL framework provides tools for specifying a task using an image or text description. This allows us to work with more complex models of shoebox tasks and also makes the application more intuitive. In robot-assisted teaching, teachers should have a choice of how to work with the robots: some would prefer to use only images, others would be interested in advanced methods that allow to describe difficult tasks with many objects. 

\begin{figure}
    \centering
    \includegraphics[width=1\linewidth]{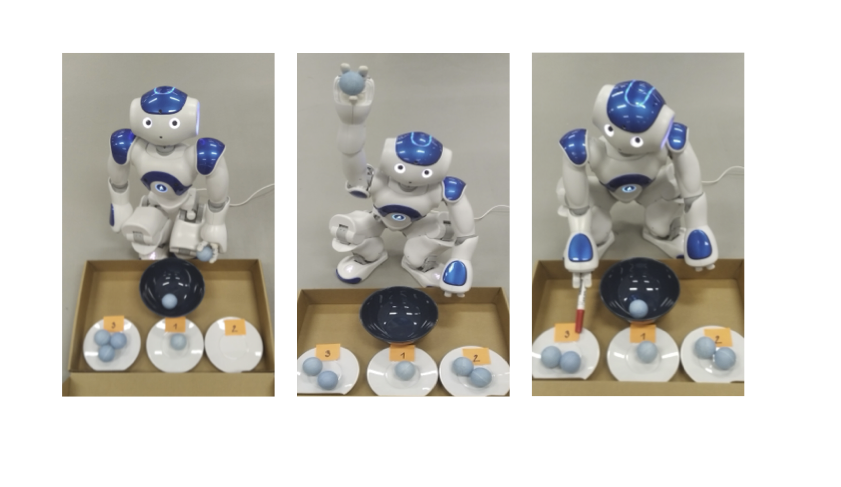}
    \caption{Examples of converting a planned action into a real robot action according to the selected teaching strategy: moving the object to a target location, passing the object to the student and pointing to the object.}
    \label{fig:method}
\end{figure}

\subsubsection{Collaboration with experts in ASD}

In the preparation phase, we will discuss basic practical questions related to structure teaching, e.g., which structured tasks types are currently used in education (in Czech Republic and in world), are there any standard for shoe-box tasks in Czech schools (for example, products from ShoeboxTasks®) or are they typically designed and made by teachers, what are physical parameters of the boxes.

We also need to consult educational strategies and methodology specific to ASD, especially the safety recommendations for experiments with students and the real robot. For testing, annotated data from the multi-modal database of autistic children’s interactions are
available~\cite{lada21}.

\section{Conclusion}

In this work, we introduced Image2PDDL, a framework that leverages VLM to automate the generation of PDDL problems. By utilizing both images and textual descriptions for initial and goal states, Image2PDDL bridges the gap between visual perception and symbolic planning, allowing for a streamlined, accessible process of problem generation across diverse domains. Our evaluation across three domains—Blocksworld, Sliding-Tile Puzzle, and Kitchen—demonstrated that Image2PDDL achieves high syntax correctness, reliably producing grammatically valid PDDL outputs across various task complexities and input types. While content correctness varied depending on domain difficulty and input modality, results indicated that the framework performs well with structured tasks and benefits from textual input in visually complex scenarios.

Image2PDDL’s promising performance suggests broader applications in AI planning, particularly for complex real-world tasks requiring visual and symbolic integration. Future work could enhance the framework’s capacity to manage intricate object relationships and multi-step transformations in dynamic environments, such as the autistic shoe-box task. Overall, this framework lays the groundwork for more accessible and scalable AI planning solutions, creating new opportunities for integrating vision and language models into automated planning domains.

We discussed an apparent application case for Image2PDDL in robot-assisted teaching of students with ASD. Some possible research avenues include
\begin{itemize}
\item automatically generating 3D models of shoebox tasks 
(to provide the visual feedback or to
practice the task in a simulation).
\item 
using vision language model to generate verbal instructions.
\item
automating the entire process: use the robot to prepare the shoebox task or the entire work session.
\item 
using the system as tool for education of robot programming for students with ASD.
\end{itemize}

\section{Acknowledgements}
The presented work has been supported by the Czech Science Foundation (GA\v{C}R) under the research project number 22-30043S.

\bigskip

\bibliography{aaai25}

\end{document}